\PassOptionsToPackage{table, x11names}{xcolor}
\documentclass[11pt,a4paper]{article}
\usepackage[utf8]{inputenc}
\usepackage[hyperref]{acl2020}
\usepackage{times}
\usepackage{latexsym}

\usepackage{microtype}

\usepackage{tabulary}
\usepackage{hhline}
\usepackage{amsthm}
\usepackage{mathtools}
\DeclarePairedDelimiterX{\infdivx}[2]{(}{)}{%
  #1\;\|\delimsize\|\;#2%
}
\newcommand{\infdiv}{D_{KL}\infdivx}
\usepackage{cleveref}
\crefformat{equation}{#2equation~#1#3}
\Crefformat{equation}{#2Equation~#1#3}
\crefformat{section}{#2section~#1#3}
\Crefformat{section}{#2Section~#1#3}
\crefformat{figure}{#2figure~#1#3}
\Crefformat{figure}{#2Figure~#1#3}
\crefformat{xnumi}{#2example~#1#3}
\Crefformat{xnumi}{#2Example~#1#3}
\crefformat{table}{#2table~#1#3}
\Crefformat{table}{#2Table~#1#3}

\DeclareUnicodeCharacter{2212}{-}

\theoremstyle{definition}
\newtheorem{definition}{Definition}[section]

\aclfinalcopy 

\title{Linguistic Features for Readability Assessment}

\author{Tovly Deutsch\qquad  Masoud Jasbi\qquad  Stuart Shieber \\ Harvard University \\ \texttt{tdeutsch@college.harvard.edu, masoud\_jasbi@fas.harvard.edu}\\\texttt{ shieber@seas.harvard.edu}\\
}

\begin{document}
\maketitle
\begin{abstract}
Readability assessment aims to automatically classify text by the level appropriate for learning readers. Traditional approaches to this task utilize a variety of linguistically motivated features paired with simple machine learning models. More recent methods have improved performance by discarding these features and utilizing deep learning models. However, it is unknown whether augmenting deep learning models with linguistically motivated features would improve performance further. This paper combines these two approaches with the goal of improving overall model performance and addressing this question. Evaluating on two large readability corpora, we find that, given sufficient training data, augmenting deep learning models with linguistically motivated features does not improve state-of-the-art performance. Our results provide preliminary evidence for the hypothesis that the state-of-the-art deep learning models represent linguistic features of the text related to readability. Future research on the nature of representations formed in these models can shed light on the learned features and their relations to linguistically motivated ones hypothesized in traditional approaches.
\end{abstract}

\section{Introduction}

Readability assessment poses the task of identifying the appropriate reading level for text. Such labeling is useful for a variety of groups including learning readers and second language learners. Readability assessment systems generally involve analyzing a corpus of documents labeled by editors and authors for reader level. Traditionally, these documents are transformed into a number of linguistic features that are fed into simple models like SVMs and MLPs \citep{schwarmReadingLevelAssessment2005, vajjalaImprovingAccuracyReadability2012}.

More recently, readability assessment models utilize deep neural networks and attention mechanisms \citep{martincSupervisedUnsupervisedNeural2019}. While such models achieve state-of-the-art performance on readability assessment corpora, they struggle to generalize across corpora and fail to achieve perfect classification. Often, model performance is improved by gathering additional data. However, readability annotations are time-consuming and expensive given lengthy documents and the need for qualified annotators. A different approach to improving model performance involves fusing the traditional and modern paradigms of linguistic features and deep learning. By incorporating the inductive bias provided by linguistic features into deep learning models, we may be able to reduce the limitations posed by the small size of readability datasets.

In this paper, we evaluate the joint use of linguistic features and deep learning models. We achieve this fusion by simply taking the output of deep learning models as features themselves. Then, these outputs are joined with linguistic features to be further fed into some other model like an SVM. We select linguistic features based on a broad psycholinguistically-motivated composition by \citet{vajjalabalakrishnaAnalyzingTextComplexity2015}. Transformers and Hierarchical attention networks were selected as the deep learning models because of their state-of-art performance in readability assessment. Models were evaluated on two of the largest available corpora for readability assessment: WeeBit and Newsela. We also evaluate with different sized training sets to investigate the use of linguistic features in data-poor contexts. Our results find that, given sufficient training data, the linguistic features do not provide a substantial benefit over deep learning methods.

The rest of this paper is organized as follows. Related research is described in \cref{sec:related}. \Cref{sec:methodology} details our preprocessing, features, and model construction. \Cref{sec:results} presents model evaluations on two corpora. \Cref{sec:conclusion} discusses the implications of our results.

We provide a publicly available version of the code used for our experiments.\footnote{\url{https://github.com/TovlyDeutsch/Linguistic-Features-for-Readability}} 

\section{Related Work}
\label{sec:related}

Work on readability assessment has involved progress on three core components: corpora, features, and models. While early work utilized small corpora, limited feature sets, and simple models, modern research has experimented with a broad set of features and deep learning techniques.

Labeled corpora can be difficult to assemble given the time and qualifications needed to assign a text a readability level. The size of readability corpora expanded significantly with the introduction of the WeeklyReader corpus by \citet{schwarmReadingLevelAssessment2005}. Composed of articles from an educational magazine, the WeeklyReader corpus contains roughly 2,400 articles. The WeeklyReader corpus was then built upon by \citet{vajjalaImprovingAccuracyReadability2012} by adding data from the BBC Bitesize website to form the WeeBit corpus. This WeeBit corpus is larger, containing roughly 6,000 documents, while also spanning a greater range of readability levels. Within these corpora, topic and readability are highly correlated. Thus, \citet{xiaTextReadabilityAssessment2016} constructed the Newsela corpus in which each article is represented at multiple reading levels thereby diminishing this correlation.  


Early work on readability assessment, such as that of \citet{fleschNewReadabilityYardstick1948}, extracted simple textual features like character count. More recently, \citet{schwarmReadingLevelAssessment2005} analyzed a broader set of features including out-of-vocabulary scores and syntactic features such as average parse tree height. \citet{vajjalaImprovingAccuracyReadability2012} assembled perhaps the broadest class of features. They incorporated measures shown by \citet{luAutomaticAnalysisSyntactic2010} to correlate well with second language acquisition measures, as well as psycholinguistically relevant features from the Celex Lexical database and MRC Psycholinguistic Database \citep{Baayen1995TheCL, wilsonMRCPsycholinguisticDatabase1988}.

Traditional feature formulas, like the Flesch formula, relied on linear models. Later work progressed to more complex related models like SVMs \citep{schwarmReadingLevelAssessment2005}. Most recently, state-of-art-performance has been achieved on readability assessment with deep neural network incorporating attention mechanisms. These approaches ignore linguistic features entirely and instead feed the raw embeddings of input words, relying on the model itself to extract any relevant features. Specifically, \citet{martincSupervisedUnsupervisedNeural2019} found that a pretrained transformer model achieved state-of-the-art performance on the WeeBit corpus while a hierarchical attention network (HAN) achieved state-of-the-art performance on the Newsela corpus.


Deep learning approaches generally exclude any specific linguistic features. In general, a ``feature-less" approach is sensible given the hypothesis that, with enough data, training, and model complexity, a model should learn any linguistic features that researchers might attempt to precompute. However, precomputed linguistic features may be useful in data-poor contexts where data acquisition is expensive and error-prone.  
For this reason, in this paper we attempt to incorporate linguistic features with deep learning methods in order to improve readability assessment.

\section{Methodology}
\label{sec:methodology}
\subsection{Corpora}
\label{sec:methodCorpora}
\subsubsection{WeeBit}
The WeeBit corpus was assembled by \citet{vajjalaImprovingAccuracyReadability2012} by combining documents from the WeeklyReader educational magazine and the BBC Bitesize educational website. They selected classes to assemble a broad range of readability levels intended for readers aged 7 to 16. To avoid classification bias, they undersampled classes in order to equalize the number of documents in each class to 625. We term this downsampled corpus ``WeeBit downsampled". Following the methodologies of \citet{xiaTextReadabilityAssessment2016} and \citet{ martincSupervisedUnsupervisedNeural2019}, we applied additional preprocessing to the WeeBit corpus in order to remove extraneous material.

\subsubsection{Newsela}
\label{sec:NewselaMethod}
 The Newsela corpus \citep{xiaTextReadabilityAssessment2016} consists of 1,911 news articles each re-written up to 4 times in simplified manners for readers at different reading levels. This simplification process means that, for any given topic, there exist examples of material on that topic suited for multiple reading levels. This overlap in topic should make the corpus more challenging to label than the WeeBit corpus. In a similar manner to the WeeBit corpus, the Newsela corpus is labeled with grade levels ranging from grade 2 to grade 12. As with WeeBit, these labels can either be treated as classes or transformed into numeric labels for regression.

\subsubsection{Labeling Approaches}
Often, readability classes within a corpus are treated as unrelated. These approaches use raw labels as distinct unordered classes. However, readability labels are ordinal, ranging from lower to higher readability. Some work has addressed this issue such as the readability models of \citet{flor-etal-2013-lexical} which predict grade levels via linear regression. To test different approaches to acknowledging this ordinality, we devised three methods for labeling the documents: ``classification", ``age regression", and ``ordered class regression".

The classification approach uses the classes originally given. This approach does not suppose any ordinality of the classes. Avoiding such ordinality may be desirable for the sake of simplicity. 

``Age regression" applies the mean of the age ranges given by the constituent datasets. For instance, in this approach Level 2 documents from Weekly Reader would be given the label of 7.5 as they are intended for readers of ages 7-8. The advantage of age regression over standard classification is that it provides more precise information about the magnitude of readability differences. 

Finally, ``ordered class regression" assigns the classes equidistant integers ordered by difficulty. The least difficult class would be labeled ``0", the second least difficult class would be labeled ``1" and so on. As with age regression, this labeling results in a regression rather than classification problem. This method retains the advantage of age regression in demonstrating ordinality. However, ordered regression labeling removes information about the relative differences in difficulty between the classes, instead asserting that they are equidistant in difficulty. The motivation behind this loss of information is that such age differences between classes may not directly translate into differences of difficulty. For instance, the readability difference between documents intended for 7 or 8 year-olds may be much greater than between documents intended for 15 or 16 year-olds because reading development is likely accelerated in younger years.

For final model inferences, we used the classification approach for comparison to previous work. For intermediary CNN models, all three approaches were tested. As the different approaches with CNN models produced insubstantial differences, other model types were restricted to the simple classification approach.

\subsection{Features}
\label{sec:methodLingFeatures}
Motivated by the success in using linguistic features for modeling readability, we considered a large range of textual analyses relevant to readability. In addition to utilizing features posed in the existing readability research, we investigated formulating new features with a focus on syntactic ambiguity and syntactic diversity. This challenging aspect of language appeared to be underutilized in existing readability literature.
\subsubsection{Existing Features}
To capture a variety of features, we utilized existing linguistic feature computation software\footnote{This code can be found at \url{https://bitbucket.org/nishkalavallabhi/complexity-features}.} developed by \citet{vajjalabalakrishnaAnalyzingTextComplexity2015} based on 86 feature descriptions in existing readability literature. Given the large number of features, in this section we will focus on the categories of features and their psycholinguistic motivations (where available) and properties. The full list of features used can be found in appendix A.

\paragraph{Traditional Features}

The most basic features involve what  \citet{vajjalaImprovingAccuracyReadability2012} refer to as ``traditional features" for their use in long-standing readability formulae. They include characters per word, syllables per word, and traditional formulas based on such features like the Flesch-Kincaid formula \citep{kincaidDerivationNewReadability1975}.

Another set of feature types consists of counts and ratios of part-of-speech tags, extracted using the Stanford parser \citep{kleinAccurateUnlexicalizedParsing2003}. In addition to basic parts of speech like nouns, some features include phrase level constituent counts like noun phrases and verb phrases. All of these counts are normalized by either the number of word tokens or number of sentences to make them comparable across documents of differing lengths. These counts are not provided with any psycholinguistic motivation for their use; however, it is not an unreasonable hypothesis that the relative usage of these constituents varies across reading levels. Empirically, these features were shown to have some predictive power for readability. In addition to parts of speech counts, we also utilized word type counts as a simple baseline feature, that is, counting the number of instances of each possible word in the vocabulary. These counts are also divided by document length to generate proportions.

Becoming more abstract than parts of speech, some features count complex syntactic constituent like clauses and subordinated clauses. Specifically, \citet{luAutomaticAnalysisSyntactic2010} found ratios involving sentences, clauses, and t-units\footnote{Defined by \citet{vajjalaImprovingAccuracyReadability2012} to be ``one main clause plus any subordinate clause or non-clausal structure that is attached to or embedded in it".} that correlated with second language learners' abilities to read a document. For many of the multi-word syntactic constituents previously described, such as noun phrases and clauses, features were also constructed of their mean lengths. Finally, properties of the syntactic trees themselves were analyzed such as their mean heights.

 Moving beyond basic features from syntactic parses, \citet{vajjalabalakrishnaAnalyzingTextComplexity2015} also incorporated ``word characteristic" features from linguistic databases. A significant source was the Celex Lexical Database \citet{Baayen1995TheCL} which ``consists of information on the orthography, phonology, morphology, syntax and
frequency for more than 50,000 English lemmas". The database appears to have a focus on morphological data such as whether a word may be considered a loan word and whether it contains affixes. It also contains syntactic properties that may not be apparent from a syntactic parse, e.g. whether a noun is countable. The MRC Psycholinguistic Database \citet{wilsonMRCPsycholinguisticDatabase1988} was also used with a focus on its age of acquisition ratings for words, an clear indicator of the appropriateness of a document's vocabulary.

\subsubsection{Novel Syntactic Features}

We investigated additional syntactic features that may be relevant for readability but whose qualities were not targeted by existing features. These features were used in tandem with the existing linguistic features described previously; future work could utilize these novel feature independently to investigate their particular effect on readability information extraction.  For generating syntactic parses, we used the PCFG (probabilistic context-free grammar) parser \citep{kleinAccurateUnlexicalizedParsing2003} from the Stanford Parser package.

\paragraph{Syntactic Ambiguity}

Sentences can have multiple grammatical syntactic parses. Therefore, syntactic parsers produce multiple parses annotated with parse likelihood. It may seem sensible to use the number of parses generated as a measure of ambiguity. However, this measure is extremely sensitive to sentence length as longer sentences tend to have more possible syntactic parses. Instead, if this list of probabilities is viewed as a distribution, the standard deviation of this distribution is likely to correlate with perceptions of syntactic ambiguity. 

\theoremstyle{definition}
\begin{definition}{$PD_x$}

The parse deviation, $PD_x(s)$, of sentence $s$ is the standard deviation of the distribution of the $x$ most probable parse log probabilities for $s$. If $s$ has less than $x$ valid parses, the distribution is taken from all the valid parses.
\end{definition}

For large values of $x$, $PD_x(s)$ can be significantly sensitive to sentence length: longer sentences are likely to have more valid syntactic parses and thus create low probability tails that increase standard deviation. To reduce this sensitivity, an alternative involves measuring the difference between the largest and mean parse probability.

\theoremstyle{definition}
\begin{definition}{$PDM_x$}

$PDM_x(s)$ is the difference between the largest parse log probability and the mean of the log probabilities of the $x$ most probable parses for a sentence s. If $s$ has less than $x$ valid parses, the mean is taken over all the valid parses.
\end{definition}

As a compromise between parse investigation and the noise of implausible parses, we selected $PDM_{10}$, $PD_{10}$, and $PD_2$ as features to use in the models of this paper. 

\paragraph{Part-of-Speech Divergence}

To capture the grammatical makeup of a sentence or document, we can count the usage of each part of speech (``POS"), phrase, or clause. The counts can be collected into a distribution. Then, the standard deviation of this distribution, $POSD_{dev}$, measures a sentence's grammatical heterogeneity.

\theoremstyle{definition}
\begin{definition}{$POSD_{dev}$}

$POSD_{dev}(d)$ is the standard deviation of the distribution of POS counts for document $d$.
\end{definition}

Similarly, we may want to measure how this grammatical makeup differs from the composition of the document as a whole, a concept that might be termed syntactic uniqueness. To capture this concept, we measure the Kullback-Leibler divergence \citep{kullbackInformationSufficiency1951} between the sentence POS count distribution and the document POS count distribution.

\theoremstyle{definition}
\begin{definition}{$POS_{div}$}

Let $P(s)$ be the distribution of POS counts for sentence $s$ in document $d$.
Let $Q$ be the distribution of POS counts for document $d$.
Let $|d|$ be the number of sentences in $d$.
$$POS_{div}(d) = \sum_{s \in d} \frac{\infdiv{P(s)}{Q}}{|d|}$$
\end{definition}

\subsection{Models}

A large range of model complexities were evaluated in order to ascertain the performance improvements, or lack thereof, of additional model complexity. In this section we will describe the specific construction and usage of these models for the experiments conducted in this paper, ordered roughly by model complexity.

\paragraph{SVMs, Linear Models, and Logistic Regression}

We used the Scikit-Learn library \citep{pedregosaScikitlearnMachineLearning2011} for constructing SVM models. Hyper-parameter optimization was performed using the guidelines suggested by \citet{hsuPracticalGuideSupport2003}. From the Scikit-Learn library, we also utilized the linear support vector classifier (an SVM with a linear kernel) and logistic regression classifier. As simplicity was the aim for these evaluations, no hyperparameter optimization was performed. The logistic regression classifier was trained using the stochastic average gradient descent (``sag") optimizer.

\paragraph{CNN}
Convolutional neural networks were selected for their demonstrated performance on sentence classification \citep{kimConvolutionalNeuralNetworks2014}. The CNN model used in this paper is based on the one described by \citet{kimConvolutionalNeuralNetworks2014} and implemented using the Keras \citep{chollet2015keras}, Tensorflow \citep{tensorflow2015-whitepaper}, and Magpie libraries.

\paragraph{Transformer}
\label{sec:methodTransformer}
The transformer \citep{vaswaniAttentionAllYou2017} is a neural-network-based model that has achieved state-of-the-art results on a wide array of natural language tasks including readability assessment \citep{martincSupervisedUnsupervisedNeural2019}. Transformers utilize the mechanism of attention which allows the model to attend to specific parts of the input when constructing the output. Although they are formulated as sequence-to-sequence models, they can be modified to complete a variety of NLP tasks by placing an additional linear layer at the end of the network and training that layer to produce the desired output. This approach often achieves state-of-the-art results when combined with pretraining. In this paper, we use the BERT \citep{devlinBERTPretrainingDeep2019} transformer-based model that is pretrained on BooksCorpus (800M words) \citep{zhuAligningBooksMovies2015} and English Wikipedia. The model is then fine-tuned on a specific readability corpus such as WeeBit. The pretrained BERT model is sourced from the Huggingface transformers library \citep{Wolf2019HuggingFacesTS} and is composed of 12 hidden layers each of size 768 and 12 self-attention heads. The fine-tuning step utilizes an implementation by \citet{martincSupervisedUnsupervisedNeural2019}. Among the pretrained transformers in the Huggingface library, there are transformers that can accept sequences of size 128, 256, and 512. The 128 sized model was chosen based on the finding by \citet{martincSupervisedUnsupervisedNeural2019} that it achieved the highest performance on the WeeBit and Newsela corpora. Documents that exceeded the input sequence size were truncated.

\paragraph{HAN}
The Hierarchical attention network involves feeding the input through two bidirectional RNNs each accompanied by a separate attention mechanism. One attention mechanism attends to the different words within each sentence while the second mechanism attends to the sentences within the document. These hierarchical attention mechanisms are thought to better mimic the structure of documents and consequently produce superior classification results. The implementation of the model used in this paper is identical to the original architecture described by \citet{yangHierarchicalAttentionNetworks2016} and was provided by the authors of \citet{martincSupervisedUnsupervisedNeural2019} based on code by \citet{nguyenHierarchicalAttentionNetworks2020}.
\label{sec:methodHan}

\subsection{Incorporating Linguistic Features with Neural Models}
\label{lingPlusNeural}
The neural network models thus far described take either the raw text or word vector embeddings of the text as input. They make no use of linguistic features such as those described in \cref{sec:methodLingFeatures}. We hypothesized that combining these linguistic features with the deep neural models may improve their performance on readability assessment. Although these models theoretically represent similar features to those prescribed by the linguistic features, we hypothesized that the amount of data and model complexity may be insufficient to capture them. This can be evidenced in certain models failure to generalize across readability corpora. \citet{martincSupervisedUnsupervisedNeural2019} found that the BERT model performed well on the WeeBit corpus, achieving a weighted F1 score of 0.8401, but performed poorly on the Newsela corpus only achieving an F1 score of 0.5759. They posit that this disparity occurred ``because BERT is pretrained as a language model, [therefore] it tends
to rely more on semantic than structural differences during the classification phase
and therefore performs better on problems with distinct semantic differences between
readability classes". Similarly a HAN was able to achieve better performance than BERT on the Newsela but performed substantially worse on the WeeBit corpus. Thus, under some evaluations the models have deficiencies and fail to generalize. Given these deficiencies, we hypothesized that the inductive bias provided by linguistic features may improve generalizability and overall model performance.

In order to weave together the linguistic features and neural models, we take the simple approach of using the single numerical output of a neural model as a feature itself, joined with linguistic features, and then fed into one of the simpler non-neural models such as SVMs. SVMs were chosen as the final classification model for their simplicity and frequent use in integrating numerical features. The output of the neural model could be any of the label approaches such as grade classes or age regressions described in \cref{sec:methodCorpora}. While all these labeling approaches were tested for CNNs, insubstantial differences in final inferences led us to restrict intermediary results to simple classification for other model types.

\subsection{Training and Evaluation Details}
All experiments involved 5-fold cross validation. All neural-network-based models were trained with the Adam optimizer \citep{DBLP:journals/corr/KingmaB14} with learning rates of $10^{-3}$,$10^{-4}$, and $2^{-5}$ for the CNN, HAN, and transformer respectively. The HAN and CNN models were trained for 20 and 30 epochs. The transformer models were fine-tuned for 3 epochs.

All results are reported as either a weighted F1 or macro F1 score. To calculate weighted F1, first the F1 score is calculated for each class independently, as if each class was a case of binary classification. Then, these F1 score are combined in a weighted mean in which each class is weighted by the number of samples in that class. Thus, the weighted F1 score treats each sample equally but prioritizes the most common classes. The macro F1 is similar to the weighted F1 score in that F1 scores are first calculated for each class independently. However, for the macro F1 score, the class F1 scores are combined in a mean without any weighting. Therefore, the macro F1 score treats each class equally but does not treat each sample equally, deprioritizing samples from large classes and prioritizing samples from small classes.

\section{Results}
\label{sec:results}
In this section we report the experimental results of incorporating linguistic features into readability assessment models. The two corpora, WeeBit and Newsela, are analyzed individually and then compared. Our results demonstrate that, given sufficient data, linguistic features provide little to no benefit compared to independent deep learning models. While the corpus experiment results demonstrate a portion of the approaches tested, the full results are available in appendix B

\subsection{Newsela Experiments}
\label{sec:NewselaExperiments}

For the Newsela corpus, while linguistic features were able to improve the performance of some models, the top performers did not utilize linguistic features. The results from the top performing models are presented in \cref{tab:NewselaBest}.

\definecolor{han}{rgb}{0.88,1.0,1.0}
\definecolor{transformer}{rgb}{1.0,1.0,0.88}
\definecolor{nondeep}{rgb}{0.96,0.96,0.96}
\definecolor{cnn}{RGB}{255, 230, 180}

\begin{table}
\begin{tabular}{p{5cm}c}
Features & Weighted F1\\
\hline
\rowcolor{han} HAN & 0.8024 \\
\rowcolor{han} SVM with HAN and linguistic features & 0.8014 \\
\rowcolor{han} SVM with HAN & 0.7931 \\
\rowcolor{nondeep} SVM with linguistic features and Flesch Features & 0.7694 \\
\rowcolor{transformer} SVM with transformer and linguistic features & 0.7678 \\
\rowcolor{transformer} SVM with transformer, Flesch features, and linguistic features & 0.7627 \\
\rowcolor{nondeep} SVM with Linguistic features & 0.7582 \\
\rowcolor{cnn} SVM with CNN age regression and linguistic features & 0.7281 \\
\rowcolor{cnn} SVM with CNN ordered classes regression and linguistic features & 0.7231 \\
\rowcolor{transformer} SVM with transformer and Flesch features & 0.7186 \\\hline\hline
\rowcolor{transformer} Transformer & 0.5435 \\
\rowcolor{cnn} CNN & 0.3379 \\
\end{tabular}
    \caption{Top 10 performing model results, transformer, and CNN on the Newsela corpus}
    \label{tab:NewselaBest}
\end{table}

While the HAN performance was not surpassed by models with linguistic features, the transformer models were. This improvement indicates that linguistic features capture readability information that transformers cannot capture or have insufficient data to learn. The outsize effect of adding the linguistic features to the transformer models, resulting in a weighted F1 score improvement of 0.22, may reveal what types of information they address. \citet{martincSupervisedUnsupervisedNeural2019} hypothesize that a pretrained language model ``tends to rely more on semantic than structural differences" indicating that these features are especially suited to providing non-semantic information such as syntactic qualities.

\subsection{WeeBit Experiments}

The WeeBit corpus was analyzed in two perspectives: the downsampled dataset and the full dataset. Raw results and model rankings were largely comparable between the two dataset sizes. 

\subsubsection{Downsampled WeeBit Experiments}

As with the Newsela corpus, the downsampled WeeBit corpus demonstrates no gains from being analyzed with linguistic features. The best performing model, a transformer, did not utilize linguistic features. The results for some of the best performing models are shown in \cref{tab:weebitBest}. 

\begin{table}
\begin{tabular}{p{5cm}c}
Features & Weighted F1\\
\hline
\rowcolor{transformer} Transformer & 0.8387 \\
\rowcolor{transformer} SVM with transformer, Flesch features, and linguistic features & 0.8381 \\
\rowcolor{transformer} SVM with transformer and Flesch features & 0.8359 \\
\rowcolor{transformer} SVM with transformer and linguistic features & 0.8344 \\
\rowcolor{transformer} SVM with transformer & 0.8343 \\
\rowcolor{nondeep} Logistic regression classifier with word types, Flesch features, and linguistic features & 0.8135 \\
\rowcolor{nondeep} Logistic regression classifier with word types & 0.7894 \\
\rowcolor{nondeep} Logistic regression classifier with word types, word count, and Flesch features & 0.7934 \\
\rowcolor{cnn} SVM with CNN classifier and linguistic features & 0.7923 \\
\rowcolor{nondeep} Logistic regression classifier with word types and word count & 0.7908 \\\hline\hline
\rowcolor{cnn} CNN & 0.7859 \\
\rowcolor{han} HAN & 0.7507 \\
\end{tabular}
    \caption{Top 10 performing model results, CNN, and HAN on the downsampled WeeBit corpus}
    \label{tab:weebitBest}
\end{table}

Differing with the Newsela corpus, the word type models performed near the top results on the WeeBit corpus comparably to the transformer models. Word type models have no access to word order, thus semantic and topic analysis form their core analysis. Therefore, this result supports the hypothesis of \citet{martincSupervisedUnsupervisedNeural2019} that the pretrained transformer is especially attentive to semantic content. This result also indicates that the word type features can provide a significant portion of the information needed for successful readability assessment.

The differing best performing model types between the two corpora are likely due to differing compositions. Unlike the Newsela corpus, the WeeBit corpus shows strong correlation between topic and difficulty. Extracting this topic and semantic content is thought to be a particular strength of the transformer \citep{martincSupervisedUnsupervisedNeural2019}
leading to its improved results on this corpus. 

\subsubsection{Full WeeBit Experiments}

All of the models were also tested on the full imbalanced WeeBit corpus, the top performing results of which are shown in \cref{tab:weebitFullBest}. Most performance figures increased modestly. However, these gains may not be seen if documents do not match the distribution of this imbalanced dataset. Additionally, the ranking of models between the downsampled and standard WeeBit corpora showed little change. Although the SVM with transformer and linguistic features performed better than the transformer alone, this difference is extremely small ($<0.005$) and thus not likely to be statistically significant.

\begin{table}
\rowcolors{2}{gray!25}{white}
\begin{tabular}{p{5cm}c}
\rowcolor{white}
Features & Weighted F1\\
\hline
\rowcolor{transformer} SVM with transformer and linguistic features & 0.8769 \\
\rowcolor{transformer} SVM with transformer and Flesch features & 0.8746 \\
\rowcolor{transformer} SVM with transformer & 0.8729 \\
\rowcolor{transformer} Transformer & 0.8721 \\
\rowcolor{transformer} SVM with transformer, Flesch features, and linguistic features & 0.8721 \\
\end{tabular}
    \caption{Top 5 performing model results on the WeeBit corpus}
    \label{tab:weebitFullBest}
\end{table}

\subsection{Effects of Training Set Size}
\begin{figure}
    \centering
    \includegraphics[width=1.0\columnwidth]{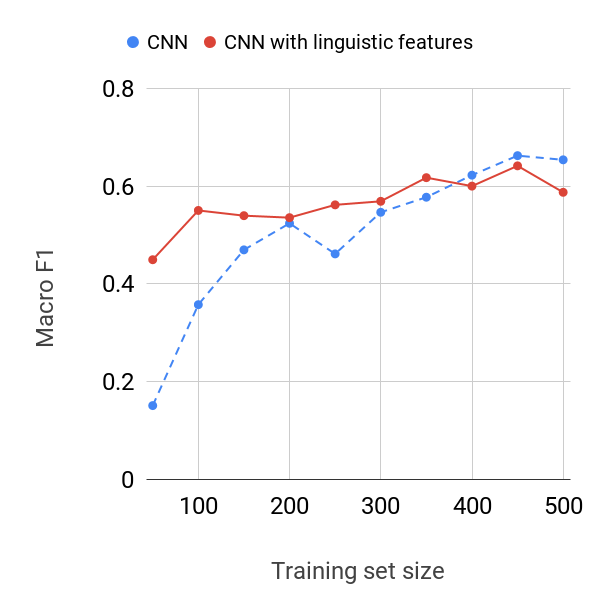}
    \caption{Performance differences across different training set sizes on the downsampled WeeBit corpus}
    \label{fig:subsampleWeeBit}
\end{figure}

One hypothesis explaining the lack of effect of linguistic features is that models learn to extract those features given enough data. Thus, perhaps in more data-poor environments the linguistic features would prove more useful. To test this hypothesis, we evaluated two CNN-based models, one with linguistic features and one without, with various sized training subsets of the downsampled WeeBit corpus. The macro F1 at these various dataset sizes is shown in \cref{fig:subsampleWeeBit}. Across the trials at different training set sizes, the test set is held constant thereby isolating the impact of training set size.

The hypothesis holds true for extremely small subsets of training data, those with fewer than 200 documents. Above this training set size, the addition of linguistic features results in insubstantial changes in performance. Thus, either the patterns exposed by the linguistic features are learnable with very little data or the patterns extracted by deep learning models differ significantly from the linguistic features. The latter appears more likely given that linguistic features are shown to improve performance for certain corpora (Newsela) and model types (transformers). 

This result indicates that the use of linguistic features should be considered for small datasets. However, the dataset size at which those features lose utility is extremely small. Therefore, collecting additional data would often be more efficient than investing the time to incorporate linguistic features.

\subsection{Effects of Linguistic Features}

Overall, the failure of linguistic features to improve state-of-the-art deep learning models indicates that, given the available corpora, model complexity, and model structures, they do not add information over and beyond what the state-of-the-art models have already learned. However, in certain data-poor contexts, they can improve the performance of deep learning models. Similarly, with more diverse and more accurately and consistently labeled corpora, the linguistic features could prove more useful. It may be the case that the best performing models already achieve near the maximal possible performance on this corpus. The reason the maximal performance may be below a perfect score (an F1 score of 1) is disagreement and inconsistency in dataset labeling. Presumably the dataset was assessed by multiple labelers who may not have always agreed with one another or even with themselves. Thus, if either a new set of human labelers or the original labelers are tasked with labeling readability in this corpus, they may only achieve performance similar to the best performance seen in these experiments. Performing this human experiment would be a useful analysis of corpus validity and consistency. Similarly, a more diverse corpus (differing in length, topic, writing style, etc.) may prove more difficult for the models to label alone without additional training data; in this case, the linguistic features may prove more helpful in providing inductive bias.

Additionally, the lack of improvement from adding linguistic features indicates that deep learning models may already be representing those features. Future work could probe the models for different aspects of the linguistic features, thereby investigating what properties are most relevant for readability.

\section{Conclusion}
\label{sec:conclusion}
In this paper we explored the role of linguistic features in deep learning methods for readability assessment, and asked: can incorporating linguistic features improve state-of-the-art models? We constructed linguistic features focused on syntactic properties ignored by existing features. We incorporated these features into a variety of model types, both those commonly used in readability research and more modern deep learning methods. We evaluated these models on two distinct corpora that posed different challenges for readability assessment. Additional evaluations were performed with various training set sizes to explore the inductive bias provided by linguistic features. While linguistic features occasionally improved model performance, particularly at small training set sizes, these models did not achieve state-of-the-art performance.

Given that linguistic features did not generally improve deep learning models, these models may be already implicitly capturing the features that are useful for readability assessment. Thus, future work should investigate to what degree the models represent linguistic features, perhaps via probing methods.

Although this work supports disusing linguistic features in readability assessment, this assertion is limited by available corpora. Specifically, ambiguity in the corpora construction methodology limits our ability to measure label consistency and validity. Therefore, the maximal possible performance may already be achieved by state-of-the-art models. Thus, future work should explore constructing and evaluating readability corpora with rigorous consistent methodology; such corpora may be assessed most effectively using linguistic features. For instance, accuracy could be improved by averaging across multiple labelers.

Overall, linguistic features do not appear to be useful for readability assessment. While often used in traditional readability assessment models, these features generally fail to improve the performance of deep learning methods. Thus, this paper provides a starting point to understanding the qualities and abilities of deep learning models in comparison to linguistic features. Through this comparison, we can analyze what types of information these models are well-suited to learning.

\bibliography{LibraryBibtex}
\bibliographystyle{acl_natbib}

\onecolumn
\appendix
\section{Feature Definitions}
\label{app:appendixFeatures}

For the following definitions, if the a ratio is undefined (i.e. the denominator is zero) the result is treated as zero. \citet{vajjalaImprovingAccuracyReadability2012} define complex nominals to be: ``a) nouns plus adjective, possessive, prepositional phrase, relative clause, participle or appositive, b) nominal clauses, c) gerunds and infinitives in subject positions." Here polysyllabic means more than two syllables and ``long words" means a word with seven or more characters. Descriptions of the norms of age of acquisition ratings can be found in \citet{kupermanAgeofacquisitionRatings302012}.

\begin{table}[h]
\centering
\begin{tabular}{|c|p{13cm}|}
\hline
\textbf{Feature Name} & \textbf{Definition} \\\hline
$PD_x(s)$ & The parse deviation, $PD_x(s)$, of sentence $s$ is the standard deviation of the distribution of the $x$ most probable parse log probabilities for $s$. If $s$ has less than $x$ valid parses, the distribution is taken from all the valid parses. \\\hline
$PDM_x$ & $PDM_x(s)$ is the difference between the largest parse log probability and the mean of the log probabilities of the $x$ most probable parses for a sentence s. If $s$ has less than $x$ valid parses, the mean is taken over all the valid parses. \\\hline
$POSD_{dev}$ & $POSD_{dev}(d)$ is the standard deviation of the distribution of POS counts for document $d$. \\\hline
$POS_{div}$ & Let $P(s)$ be the distribution of POS counts for sentence $s$ in document $d$. Let $Q$ be the distribution of POS counts for document $d$. Let $|d|$ be the number of sentences in $d$. $POS_{div}(d) = \sum_{s \in d} \frac{\infdiv{P(s)}{Q}}{|d|}$ \\\hline
\end{tabular}
\caption{Novel syntactic feature definitions}
\end{table}

\begin{table}
\centering
\begin{tabular}{|c|p{10.5cm}|}
\hline
\textbf{Feature Name} & \textbf{Definition} \\\hline
mean t-unit lenght &number of words / number of t-units \\\hline
mean parse tree height per sentence & mean parse tree height / number of sentences \\\hline
subtrees per sentence & number of subtrees / number of sentences \\\hline
SBARs per sentence & number of SBARs / number of sentences \\\hline
NPs per sentence & number of NPs / number of sentences \\\hline
VPs per sentence & number of VPs / number of sentences \\\hline
PPs per sentence & number of PPs / number of sentences \\\hline
mean NP size & number of children of NPs / number of NPs \\\hline
mean VP size & number of children of VPs / number of VPs \\\hline
mean PP size & number of children of PPs / number of PPs \\\hline
WHPs per sentence & number of wh-phrases / number of sentences \\\hline
RRCs per sentence & number of reduced relative clauses / number of sentences \\\hline
ConjPs per sentence & number of conjunction phrases / number of sentences \\\hline
clauses per sentence & number of clauses / number of sentences \\\hline
t-units per sentence & number of t-units / number of sentences \\\hline
clauses per t-unit & number of clauses / number of t-units \\\hline
complex t-unit ratio & number of t-units that contain a dependent clause / number of t-units \\\hline
dependent clauses per clause & number of dependent clauses / number of clauses \\\hline
dependent clauses per t-unit & number of dependent clauses / number of t-units \\\hline
coordinate clauses per clause & number of coordinate clauses / number of clauses \\\hline
coordinate clauses per t-unit & number of coordinate clauses / number of t-units \\\hline
complex nominals per clauses & number of complex nominals / number of clauses \\\hline
complex nominals per t-unit & number of complex nominals / number of t-units \\\hline
VPs per t-unit & number of VP / number of t-units \\\hline
\end{tabular}
\caption{Existing syntactic-parse-based feature definitions}
\end{table}

\begin{table}
\centering
\begin{tabular}{|c|p{10cm}|}
\hline
\textbf{Feature Name} & \textbf{Definition} \\\hline
nouns per word & number of nouns / number of words \\\hline
proper nouns per word & number of proper nouns / number of words \\\hline
pronouns per word & number of pronouns / number of words \\\hline
conjuctions per word & number of conjuctions / number of words \\\hline
adjectives per word & number of adjectives / number of words \\\hline
verbs per word & number of verbs / number of words \\\hline
adverbs per word & number of adverbs / number of words \\\hline
modal verbs per word & number of modal verbs / number of words \\\hline
prepositions per word & number of prepositions / number of words \\\hline
interjections per word & number of interjections / number of words \\\hline
personal pronouns per word & number of personal pronouns / number of words \\\hline
wh-pronouns per word & number of wh-pronouns / number of words \\\hline
lexical words per word & number of lexical words / number of words \\\hline
function words per word & number of function words / number of words \\\hline
determiners per word & number of determiners / number of words \\\hline
VBs per word & number of base form verbs / number of words \\\hline
VBDs per word & number of past tense verbs / number of words \\\hline
VBGs per word & number of gerund or present participle verbs / number of words \\\hline
VBNs per word & number of past participle verbs / number of words \\\hline
VBPs per word & number of non-3rd person singular present verbs / number of words \\\hline
VBZs per word & number of 3rd person singular present verbs / number of words \\\hline
adverb variation & number of adverbs / number of lexical words \\\hline
adjective variation & number of adjectives / number of lexical words \\\hline
modal verb variation & number of adverbs and adverbs / number of lexical words \\\hline
noun variation & number of nouns / number of lexical words \\\hline
verb variation-I & number of verbs / number of unique verbs \\\hline
verb variation-II & number of verbs / number of lexical words \\\hline
squared verb variation-I & $(\text{number of verbs})^2$ / number of unique verbs \\\hline
corrected verb variation-I & number of verbs / $\sqrt{2 * \text{number of unique verbs}}$ \\\hline
\end{tabular}
\caption{Existing POS-tag-based feature definitions}
\end{table}

\begin{table}
\centering
\begin{tabular}{|c|p{10cm}|}
\hline
\textbf{Feature Name} & \textbf{Definition} \\\hline
AoA Kuperman & Mean age of acquisition of words (Kuperman database) \\\hline
AoA Kuperman lemmas & Mean age of acquisition of lemmas \\\hline
AoA Bird lemmas & Mean age of acquisition of lemmas, Bird norm \\\hline
AoA Bristol lemmas & Mean age of acquisition of lemmas, Bristol norm \\\hline
AoA Cortese and Khanna lemmas & Mean age of acquisition of lemmas, Cortese and Khanna norm \\\hline
MRC familiarity & Mean word familiarity rating \\\hline
MRC concreteness & Mean word concreteness rating \\\hline
MRC Imageability & Mean word imageability rating \\\hline
MRC Colorado Meaningfulness & mean word Colorado norms meaningfulness rating \\\hline
MRC Pavio Meaningfulness & mean word Pavio norms meaningfulness rating \\\hline
MRC AoA & Mean age of acquisition of words (MRC database) \\\hline
\end{tabular}
\caption{Existing psycholinguistic feature definitions}
\end{table}

\begin{table}
\centering
\begin{tabular}{|c|p{11cm}|}
\hline
\textbf{Feature Name} & \textbf{Definition} \\\hline
number of sentences & number of sentences \\\hline
mean sentence length & number of words / number of sentences \\\hline
number of characters & number of characters \\\hline
number of syllables & number of syllables \\\hline
Flesch-Kincaid Formula & $11.8 * \text{syllables per word} + 0.39 * \text{words per sentence} - 15.59$ \\\hline
Flesch Fomula & $206.835 - 1.015 * \text{words per sentence} - 84.6 * \text{syllables per word}$ \\\hline
Automated Readability Index & $4.71 * \text{characters per word} + 0.5 * \text{words per sentence} - 21.43 $ \\\hline
Coleman Liau Formula & $-29.5873 * \text{sentences per word} + 5.8799 * \text{characters per word} - 15.8007$ \\\hline
SMOG Formula & $1.0430 * \sqrt{30.0 * \text{polysyllabic words per sentence}} + 3.1291$ \\\hline
Fog Fomula & $(\text{words per sentence} + \text{proportion of words that are polysylabic}) * 0.4$ \\\hline
FORCAST Readability Formula & $20 - 15 * \text{monosylabic words per word}$ \\\hline
LIX Readability Formula & $\text{words per sentence} + \text{long words per} word * 100.0$ \\\hline
\end{tabular}
\caption{Existing traditional feature definitions}
\end{table}

\begin{table}
\centering
\begin{tabular}{|c|p{9.7cm}|}
\hline
\textbf{Feature Name} & \textbf{Definition} \\\hline
type token ratio & number of word types / number of word tokens \\\hline
corrected type token ratio & number of word types / $\sqrt{2 * \text{number of word tokens}}$ \\\hline
root type token ratio & number of word types / $\sqrt{\text{number of word tokens}}$ \\\hline
bilogorathmic type token ratio & $log(\text{number of word types}) / log(\text{number of word tokens})$ \\\hline
uber index & $(log(\text{number of word types}))^2 / log(\frac{\text{number of word tokens}}{\text{number of word types}})$ \\\hline
measure of textual lexical diversity (MTLD) & see McCarthy and Jarvis, 2010 \\\hline
number of senses & total number of senses across all words / number of word tokens \\\hline
hyeprnyms per word & number of hypernyms / number of word tokens \\\hline
hyponyms per word & total number of senses hyponyms / number of word tokens \\\hline
\end{tabular}
\caption{Existing traditional feature definitions}
\end{table}
\clearpage
\crefalias{chapter}{appsec}
\section{Full Model Results}
\label{app:A}
\rowcolors{2}{gray!25}{white}
\begin{table}[h]
    \centering
\makebox[\textwidth][c]{
\begin{tabular}{p{7cm}p{1.5cm}p{1.5cm}p{1.5cm}p{1.5cm}p{1.5cm}p{1.5cm}}
\rowcolor{white}
Features & Weighted F1 & Macro F1 & SD weighted F1 & SD macro F1 \\
\hline
Linear classifier with Flesch Score & 0.2147 & 0.2156 & 0.0347 & 0.0253 \\
Linear classifier with Flesch features & 0.3973 & 0.3976 & 0.0154 & 0.0087 \\
SVM with HAN & 0.5531 & 0.5499 & 0.1944 & 0.1928 \\
SVM with Flesch features & 0.5908 & 0.5905 & 0.0157 & 0.0168 \\
SVM with CNN ordered class regression & 0.6703 & 0.6700 & 0.0360 & 0.0334 \\
SVM with CNN age regression & 0.6743 & 0.6742 & 0.0339 & 0.0314 \\
Linear classifier with word types & 0.7202 & 0.7189 & 0.0063 & 0.0085 \\
SVM with CNN ordered classes regression, and linguistic features & 0.7265 & 0.7262 & 0.0326 & 0.0297 \\
Logistic regression classification with word types, Flesch features, and linguistic features & 0.7382 & 0.7376 & 0.0710 & 0.0684 \\
SVM with CNN age regression and linguistic features & 0.7384 & 0.7376 & 0.0361 & 0.0346 \\
HAN & 0.7507 & 0.7501 & 0.0306 & 0.0302 \\
SVM with linguistic features and Flesch features & 0.7664 & 0.7667 & 0.0109 & 0.0114 \\
SVM with linguistic features & 0.7665 & 0.7666 & 0.0146 & 0.0153 \\
CNN & 0.7859 & 0.7852 & 0.0171 & 0.0166 \\
SVM with HAN and linguistic features & 0.7862 & 0.7864 & 0.0631 & 0.0633 \\
SVM with CNN classifier & 0.7882 & 0.7879 & 0.0217 & 0.0195 \\
Logistic regression with word types & 0.7894 & 0.7887 & 0.0151 & 0.0202 \\
Logistic regression classification with word types and word count & 0.7908 & 0.7899 & 0.0130 & 0.0182 \\
SVM with CNN classifier and linguistic features & 0.7923 & 0.7919 & 0.0210 & 0.0193 \\
Logistic regression classification with word types, word count, and Flesch features & 0.7934 & 0.7926 & 0.0135 & 0.0187 \\
Logistic regression with word types, Flesch features, and linguistic features & 0.8135 & 0.8130 & 0.0131 & 0.0169 \\
SVM with transformer & 0.8343 & 0.8340 & 0.0131 & 0.0135 \\
SVM with transformer and linguistic features & 0.8344 & 0.8347 & 0.0106 & 0.0091 \\
SVM with transformer and Flesch features & 0.8359 & 0.8358 & 0.0151 & 0.0154 \\
SVM with transformer, Flesch features, and linguistic features & 0.8381 & 0.8377 & 0.0128 & 0.0118 \\
Transformer & 0.8387 & 0.8388 & 0.0097 & 0.0073 \\
\end{tabular}
} 
    \caption{WeeBit downsampled model results sorted by weighted F1 score}
\end{table}

\begin{table}
    \centering
\makebox[\textwidth][c]{
\begin{tabular}{p{7cm}p{1.5cm}p{1.5cm}p{1.5cm}p{1.5cm}p{1.5cm}p{1.5cm}}
\rowcolor{white}
Features & Weighted F1 & Macro F1 & SD weighted F1 & SD Macro F1 \\
\hline
Linear classifier with Flesch Score & 0.3357 & 0.1816 & 0.0243 & 0.0079 \\
SVM with HAN & 0.3625 & 0.2134 & 0.0400 & 0.0331 \\
Linear classifier with Flesch features & 0.3939 & 0.2639 & 0.0239 & 0.0305 \\
SVM with Flesch features & 0.4776 & 0.3609 & 0.0222 & 0.0190 \\
SVM with CNN age regression & 0.7279 & 0.6431 & 0.0198 & 0.0205 \\
SVM with CNN ordered class regression & 0.7316 & 0.6482 & 0.0142 & 0.0141 \\
SVM with CNN age regression and linguistic features & 0.7779 & 0.7088 & 0.0156 & 0.0194 \\
SVM with CNN ordered classes regression, and linguistic features & 0.7797 & 0.7114 & 0.0130 & 0.0120 \\
Linear classifier with word types & 0.7821 & 0.7109 & 0.0162 & 0.0127 \\
SVM with Linguistic features and Flesch features & 0.7952 & 0.7367 & 0.0121 & 0.0157 \\
SVM with Linguistic features & 0.7952 & 0.7366 & 0.0130 & 0.0164 \\
HAN & 0.8065 & 0.7435 & 0.0123 & 0.0220 \\
Logistic regression classification with word types & 0.8088 & 0.7497 & 0.0127 & 0.0152 \\
Logistic regression classification with word types and word count & 0.8088 & 0.7497 & 0.0121 & 0.0148 \\
Logistic regression classification with word types, word count, and Flesch features & 0.8098 & 0.7505 & 0.0130 & 0.0163 \\
Logistic regression classification with word types, Flesch features, and linguistic features & 0.8206 & 0.7664 & 0.0428 & 0.0500 \\
CNN & 0.8282 & 0.7748 & 0.0211 & 0.0183 \\
SVM with CNN classifier and linguistic features & 0.8286 & 0.7753 & 0.0222 & 0.0209 \\
Logistic regression classification with word types, Flesch features, and ling features & 0.8293 & 0.7760 & 0.0152 & 0.0172 \\
SVM with CNN classifier & 0.8296 & 0.7754 & 0.0163 & 0.0136 \\
SVM with HAN and linguistic features & 0.8441 & 0.7970 & 0.0643 & 0.0827 \\
SVM with transformer, Flesch features, and linguistic features & 0.8721 & 0.8273 & 0.0095 & 0.0121 \\
Transformer & 0.8721 & 0.8272 & 0.0071 & 0.0102 \\
SVM with transformer & 0.8729 & 0.8288 & 0.0064 & 0.0090 \\
SVM with transformer and Flesch features & 0.8746 & 0.8305 & 0.0054 & 0.0107 \\
SVM with transformer and linguistic features & 0.8769 & 0.8343 & 0.0077 & 0.0129
\end{tabular}
} 
    \caption{WeeBit model results sorted by weighted F1 score}
\end{table}

\begin{table}
    \centering
\makebox[\textwidth][c]{
\begin{tabular}{p{7cm}p{1.5cm}p{1.5cm}p{1.5cm}p{1.5cm}p{1.5cm}p{1.5cm}}
\rowcolor{white}
Features & Weighted F1 & Macro F1 & SD weighted F1 & SD Macro F1 \\
\hline
Linear classifier with Flesch Score & 0.1668 & 0.0915 & 0.0055 & 0.0043 \\
SVM with Flesch score & 0.2653 & 0.1860 & 0.0053 & 0.0086 \\
Logistic regression with word types & 0.2964 & 0.2030 & 0.0144 & 0.0103 \\
Logistic regression with word types and word count & 0.2969 & 0.2039 & 0.0145 & 0.0095 \\
Logistic regression with word types, word count, and Flesch features & 0.3006 & 0.2097 & 0.0139 & 0.0088 \\
Linear classifier with Flesch features & 0.3080 & 0.2060 & 0.0110 & 0.0077 \\
Logistic regression with word types, Flesch features, and linguistic features & 0.3333 & 0.2489 & 0.0118 & 0.0162 \\
Linear classifier with word types & 0.3368 & 0.2485 & 0.0089 & 0.0153 \\
CNN & 0.3379 & 0.2574 & 0.0038 & 0.0111 \\
SVM with CNN classifier & 0.3407 & 0.2616 & 0.0079 & 0.0142 \\
SVM with CNN ordered class regression & 0.5207 & 0.4454 & 0.0092 & 0.0193 \\
SVM with CNN age regression & 0.5223 & 0.4469 & 0.0149 & 0.0244 \\
SVM with transformer & 0.5430 & 0.4711 & 0.0095 & 0.0258 \\
Transformer & 0.5435 & 0.4713 & 0.0106 & 0.0264 \\
Linear classifier with linguistic features & 0.5573 & 0.4748 & 0.0053 & 0.0140 \\
SVM with CNN classifier, and linguistic features & 0.7058 & 0.5510 & 0.0079 & 0.0357 \\
SVM with Flesch features & 0.7177 & 0.6257 & 0.0079 & 0.0292 \\
SVM with transformer and Flesch features & 0.7186 & 0.6305 & 0.0074 & 0.0282 \\
SVM with CNN ordered classes regression and linguistic features & 0.7231 & 0.6053 & 0.0062 & 0.0331 \\
SVM with CNN age regression and linguistic features & 0.7281 & 0.6104 & 0.0057 & 0.0337 \\
SVM with linguistic features & 0.7582 & 0.6432 & 0.0089 & 0.0379 \\
SVM with transformer, Flesch features, and linguistic features & 0.7627 & 0.6263 & 0.0075 & 0.0301 \\
SVM with transformer and linguistic features & 0.7678 & 0.6656 & 0.0230 & 0.0385 \\
SVM with linguistic features and Flesch Features & 0.7694 & 0.6446 & 0.0060 & 0.0406 \\
SVM with HAN & 0.7931 & 0.6724 & 0.0448 & 0.0449 \\
SVM with HAN and linguistic features & 0.8014 & 0.6751 & 0.0263 & 0.0379 \\
HAN & 0.8024 & 0.6775 & 0.1116 & 0.1825 \\
\end{tabular}
} 
    \caption{Newsela model results sorted by weighted F1 score}
    \label{tab:NewselaFullResults}
\end{table}

\end{document}